\pdfoutput=1
\documentclass[11pt]{article}
\usepackage{EMNLP2023}
\usepackage{times}
\usepackage{latexsym}

\usepackage[T1]{fontenc}
\usepackage[utf8]{inputenc}
\usepackage{hyperref}       %
\usepackage{url}            %
\usepackage{booktabs}       %
\usepackage{amsfonts}       %
\usepackage{nicefrac}       %
\usepackage{microtype}      %
\usepackage{xcolor}         %
\usepackage{microtype}
\usepackage{inconsolata}

\usepackage{amssymb}
\usepackage{amsmath}
\usepackage{multirow}
\usepackage{algorithm}
\usepackage{algpseudocode}
\usepackage{subcaption}
\usepackage{pifont}

\usepackage{graphicx}
\usepackage{colortbl}
\usepackage{enumitem}

\title{VPTQ: Extreme Low-bit Vector Post-Training Quantization for \\
Large Language Models}

\author{
  Yifei Liu\textsuperscript{\ddag,\dag,*} \quad
  Jicheng Wen\textsuperscript{\dag} \quad
  Yang Wang\textsuperscript{\dag,$\diamond$}\quad 
  Shengyu Ye\textsuperscript{\ddag,\dag} \quad \AND
  Li Lyna Zhang\textsuperscript{\dag} \quad
  Ting Cao\textsuperscript{\dag} \quad
  Cheng Li\textsuperscript{\ddag} \quad
  Mao Yang\textsuperscript{\dag} \\
  \textsuperscript{\ddag}University of Science and Technology of China \\
  \textsuperscript{\dag}Microsoft \\
   \{v-liuyifei, jicheng.wen, Yang.Wang92, v-shengyuye, lzhani, \\ ting.cao, maoyang\}@microsoft.com, chengli7@ustc.edu.cn
}

\usepackage{xcolor}

\usepackage{graphicx}

\newcommand\extrafootertext[1]{%
    \bgroup
    \renewcommand\thefootnote{\fnsymbol{footnote}}%
    \renewcommand\thempfootnote{\fnsymbol{mpfootnote}}%
    \footnotetext[0]{#1}%
    \egroup
}

\usepackage{amsmath,amsfonts,bm}

\def\eqref#1{equation~\ref{#1}}

\def\1{\bm{1}}

\DeclareMathAlphabet{\mathsfit}{\encodingdefault}{\sfdefault}{m}{sl}
\SetMathAlphabet{\mathsfit}{bold}{\encodingdefault}{\sfdefault}{bx}{n}

\begin{document}
\maketitle
\vspace{-5in}
\begin{abstract}
Scaling model size significantly challenges the deployment and inference of Large Language Models (LLMs). 
Due to the redundancy in LLM weights, recent research has focused on pushing weight-only quantization to extremely low-bit (even down to 2 bits). 
It reduces memory requirements, optimizes storage costs, and decreases memory bandwidth needs during inference. 
However, due to numerical representation limitations, traditional scalar-based weight quantization struggles to achieve such extreme low-bit.
Recent research on Vector Quantization (VQ) for LLMs has demonstrated the potential for extremely low-bit model quantization by compressing vectors into indices using lookup tables.

In this paper, we introduce \textbf{Vector Post-Training Quantization (VPTQ)} for extremely low-bit quantization of LLMs. 
We use Second-Order Optimization to formulate the LLM VQ problem and guide our quantization algorithm design by solving the optimization.
We further refine the weights using Channel-Independent Second-Order Optimization for a granular VQ.
In addition, by decomposing the optimization problem, we propose a brief and effective codebook initialization algorithm. 
We also extend VPTQ to support residual and outlier quantization, which enhances model accuracy and further compresses the model.
Our experimental results show that VPTQ reduces model quantization perplexity by $0.01$-$0.34$ on LLaMA-2, $0.38$-$0.68$ on Mistral-7B, $4.41$-$7.34$ on LLaMA-3 over SOTA at 2-bit, with an average accuracy improvement of $0.79$-$1.5\%$ on LLaMA-2, $1\%$ on Mistral-7B, $11$-$22\%$ on LLaMA-3 on QA tasks on average. 
We only utilize $10.4$-$18.6\%$ of the quantization algorithm execution time, resulting in a $1.6$-$1.8\times$ increase in inference throughput compared to SOTA. 
Our code is available at \url{https://github.com/microsoft/VPTQ}.

\end{abstract}

\extrafootertext{\textsuperscript{*}Contribution during internship at Microsoft Research}
\extrafootertext{\textsuperscript{$\diamond$}Corresponding author}
\extrafootertext{This paper is the result of an open-source research project, and the majority work of the project is accomplished in April 2024.}

\section{Introduction}

Large language models (LLMs) \cite{llama2, llama3} have shown excellent performance across various complex tasks as their sizes increase. 
However, the enormous weight of LLMs poses significant challenges for efficient inference and practical deployment. 
For instance, storing the LLaMA-2 70B model weights in FP16 format requires 140GB of memory, surpassing the capacity of high-end GPUs and necessitating multi-GPU deployment.
This huge size significantly affects memory capacity and hard disk storage and requires substantial bandwidth for inference. 
Weight-only quantization is a mainstream model compression technique that effectively reduces the model's size by representing floating-point numbers with fewer bits.

\begin{table}[bt!]
\centering
\caption{LLM Quantization Algorithm Comparison. VPTQ balances all dimensions and achieves SOTA.}
\label{tab:intro_table}
\resizebox{\columnwidth}{!}{%
\begin{tabular}{|c|c|c|c|c||c|c|}
\hline
 &
  VPTQ &
  AQLM &
  QuIP\# &
  GPTVQ &
  GPTQ &
  AWQ \\ \hline
Effective Bitwidth &
  \cellcolor[HTML]{DFFFD3}\pmb{$\downarrow$} &
  \cellcolor[HTML]{DFFFD3}$\downarrow$ &
  \cellcolor[HTML]{DFFFD3}$\downarrow$ &
  \cellcolor[HTML]{FFD3DF}$\uparrow$ &
  \cellcolor[HTML]{FFD3DF}$\uparrow\uparrow$ &
  \cellcolor[HTML]{FFD3DF}$\uparrow\uparrow$ \\ \hline
Accuracy @ Low-bit &
  \cellcolor[HTML]{DFFFD3}\pmb{$\uparrow$} &
  \cellcolor[HTML]{DFFFD3}$\uparrow$ &
  \cellcolor[HTML]{DFFFD3}$\uparrow$ &
  \cellcolor[HTML]{FFD3DF}$\downarrow$ &
  \cellcolor[HTML]{FFD3DF}$\downarrow\downarrow$ &
  \cellcolor[HTML]{FFD3DF}$\downarrow\downarrow$ \\ \hline
Quantization Time Cost &
  \cellcolor[HTML]{DFFFD3}\pmb{$\downarrow$} &
  \cellcolor[HTML]{FFD3DF}$\uparrow\uparrow$ &
  \cellcolor[HTML]{DFFFD3}$\downarrow$ &
  \cellcolor[HTML]{DFFFD3}$\downarrow$ &
  \cellcolor[HTML]{DFFFD3}$\downarrow$ &
  \cellcolor[HTML]{DFFFD3}$\downarrow$ \\ \hline
Inference Throughput &
  \cellcolor[HTML]{DFFFD3}\pmb{$\uparrow$} &
  \cellcolor[HTML]{DFFFD3}$\uparrow$ &
  \cellcolor[HTML]{FFD3DF}$\downarrow$ &
  \cellcolor[HTML]{DFFFD3}$\uparrow$ &
  \cellcolor[HTML]{DFFFD3}$\uparrow$ &
  \cellcolor[HTML]{DFFFD3}$\uparrow$ \\ \hline
\end{tabular}%
}
\end{table}

In weight-only quantization of LLMs, a prominent method is Post-Training Quantization (PTQ).
PTQ quantizes model weights directly without retraining the model. Typically, PTQ only involves converting model weights into lower-bit fixed-point numbers. 
Currently, the main approach in PTQ is scalar quantization, which converts each scalar weight in the model into a lower bit value. 
Recent work \citep{gptq, awq, SmoothQuant, owq, QuIP} has achieved near-original model accuracy with $3$-$4$ bit quantization.
Table \ref{tab:intro_table} summarizes the characteristics of typical scalar quantization methods (GPTQ, AWQ)  in LLM.
However, due to the limitations of numerical representation, traditional scalar-based weight quantization struggles to achieve extremely low-bit levels. 
For instance, with 2-bit quantization, we can only use four numerical values to represent model weights, which severely limits the range of weight representation. 
Although BitNet \citep{bitnet, bitnet158} has enabled quantization-aware training that can quantize weights to below 2 bits during the model's pre-training phase, this approach requires substantial GPU cluster resources to maintain reasonable accuracy.

Recent studies \citep{GPTVQ, QuIPsharp, AQLM} have explored an efficient method of weight-only quantization known as Vector Quantization (VQ). 

VQ is a data compression technique that maps high-dimensional vectors to a set of predefined lower-dimensional vectors stored in codebooks (lookup tables). During encoding, each data point is represented by the index of a corresponding vector in the codebook, and during decoding, the original data is approximated using these indices. 
This method substantially reduces the storage requirements for data while allowing for the quick reconstruction of original vectors through simple index references.
VQ achieves more effective data compression than scalar quantization by leveraging correlations and redundancies across different data dimensions.
By detecting and leveraging interdependence, VQ can encode complex multidimensional data with fewer bits, thus achieving higher compression ratios and reduced bit width.

While Vector Quantization (VQ) shows promise in extreme low-bit weight compression for Large Language Models (LLMs), it faces several significant challenges. Table \ref{tab:intro_table} compares the strengths and weaknesses of various VQ algorithms in multiple dimensions.

\textbf{The first challenge is ensuring the accuracy after extreme low-bit VQ quantization. }
Unlike scalar quantization, the quantization granularity of VQ algorithms is vector-based. 
The quantization may introduce additional accumulation errors due to the simultaneous quantization of multiple numbers. 
For example, GPTVQ \citep{GPTVQ} uses the Second-Order Optimization method to implement PTQ. 
However, GPTVQ accumulates quantization errors within vector quantization, leading to an inevitable increase in quantization errors as the vector length increases. This prevents the use of longer vectors and, consequently, limits the compression ratio.

\textbf{The second challenge lies in efficiently executing VQ quantization on LLMs.} 
VQ can compress vectors in the weight matrix into indices, but these indices are discrete, non-differentiable integers. This introduces difficulties in implementing VQ quantization methods through model training. For instance, AQLM \citep{AQLM} employs beam search and backpropagation to quantize and update centroids in lookup tables. VQ necessitates additional gradient estimation, slowing the convergence of model quantization training and requiring intensive training efforts to achieve better accuracy.

\textbf{The third challenge arises as the dequantization overhead in VQ model inference.}
To reduce quantization errors, complex data preprocessing methods may be used to process weights. 
QuIP\# \citep{QuIPsharp} introduces incoherence processing using the randomized Hadamard transform for the weight matrix before VQ. 
These preprocessing steps can reduce quantization errors and improve model accuracy. 
However, postprocessing must be performed in real time during model inference, which can severely impact throughput in inference.

VPTQ seeks to bypass the limitations of current VQ by offering a lightweight and efficient approach exclusively for extreme low-bit weight quantization.

In this paper, we present \textbf{Vector Post-Training Quantization (VPTQ)}, a novel approach for extremely low-bit quantization of LLMs. 
\begin{enumerate}[noitemsep]
    \item VPTQ achieves SOTA accuracy results on extremely low-bit LLMs. We formulate the quantization problem as an optimization problem and employ Second-Order Optimization to guide our quantization algorithm design. By Channel-Independent Second-Order Optimization, VPTQ reduces model quantization perplexity by $0.01$-$0.34$, $4.41$-$7.34$, $0.38$-$0.5$ on LLaMA-2/3/Mistral-7B, respectively, over SOTA at 2-bit, with an accuracy improvement of $0.79$-$1.5\%$,$11$-$22\%$,$1\%$, on LLaMA-2/3/Mistral-7B in QA tasks on average. 
    
    \item VPTQ can transform LLMs into extremely low-bit models with a minor quantization algorithm overhead. Under the guidance of the optimization problem, we transform the quantization algorithm into a heuristic algorithm to solve the optimization problem. 
    We also analyze and propose a brief and effective codebook initialization algorithm to reduce the extra overhead of centroid training and updates. Experiments show that VPTQ only requires $10.4$-$18.6\%$ of the quantization algorithm execution time compared to existing SOTA results.
    
    \item VPTQ has low dequantization overhead. VPTQ algorithm quantizes all the weights in every Linear Operator in the model into an index matrix and codebooks. During model inference, we only need to dequantize the weight matrix by reading centroids from the codebook according to the index before executing the operator. The models quantized by VPTQ result in $1.6$-$1.8\times$ improvement in inference throughput compared to SOTA.
\end{enumerate}

\section{Background and Motivation}
\subsection{Post Training Quantization in LLM}
Post-Training Quantization (PTQ) \citep{OBD, OBS, obs2order, gptq, woodfisher} aims to decrease model weight size by simplifying the numerical representation and seeking to maintain the model's accuracy without retraining the model.
We can formulate PTQ as the following optimization problem:
\begin{equation*}
\begin{aligned}
    & \arg\min \quad \mathbb{E} [ \mathcal{L}(\mathbf{X}, \mathbf{W} + \Delta \mathbf{W}) - \mathcal{L}(\mathbf{X}, \mathbf{W})] \\
    & \approx \Delta \mathbf{W}^T \cdot g{(\mathbf{W})} + \frac{1}{2}\Delta \mathbf{W}^T \cdot H{(\mathbf{W})} \cdot \Delta \mathbf{W}
\end{aligned}
\label{eq:exp_optimization}
\end{equation*}
where $\mathbf{W}$ is the original model weights, $\mathbf{\hat{W}}$ is quantized weights, and $\Delta \mathbf{W} = \mathbf{\hat{W}} - \mathbf{W}$ represents the weight quantization error. 
The loss of the model task is $\mathcal{L}$.
The optimization object is to minimize the impact of model quantization on the model task, which means minimizing the expected deviation of the loss function.

PTQ typically employs a concise and accurate method for analyzing the above optimization problem: Second-Order Optimization. Following a Taylor series expansion, this method breaks down the optimization goal into first-order, second-order, and higher-order terms.
$g(\mathbf{W})$ and $H(\mathbf{W})$ represent the gradient and Hessian of task loss $\mathcal{L}$, respectively.
It often assumes that the model has already reached local optimum before model quantization, which means that the first-order term is nearly zero. 
Higher-order terms exert a minor effect on the optimization goal, and we typically disregard interactions among weights between different layers. 
Consequently, we can simplify the optimization problem by focusing on optimizing the second-order term and then define the following optimization problem:
\begin{equation}
\begin{aligned}
    \arg \min_{\Delta \mathbf{W} } \quad & \Delta \mathbf{W}^T \cdot H(\mathbf{W}) \cdot \Delta \mathbf{W}, \\ 
    & \text{s.t. \quad} \Delta \mathbf{W} = \mathbf{0}
\end{aligned} \label{eq:optimization_problem}
\end{equation}
The objective of the optimization problem is to minimize the second-order error in model quantization, subject to the constraint that the change in model weights is as minimized as possible, i.e., $\Delta \mathbf{W} = \mathbf{0}$.

\subsection{Vector Quantization in Neural Networks}
\begin{figure}[t!]
    \centering
    \includegraphics[width=\columnwidth]{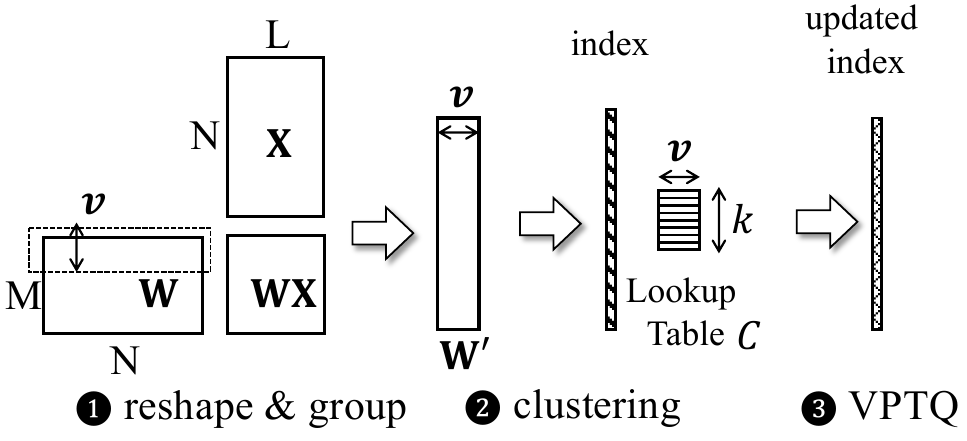}
    \caption{Vector Quantization in Weight Quantization}
    \label{fig:vq_example}
\end{figure}

VQ is a key method for efficient lossy data compression \citep{Gersho}. 
Its objective is to reduce the distortion by mapping high-dimensional original data to a lower-dimensional space represented by a lookup table (Eq. \ref{eq:vq_eq}). 
VQ maps original vectors ($\mathbf{W'}$) from the vector space to a finite set of vectors, which is commonly referred to as a codebook (lookup table, $\mathcal{C}$).
Each vector in the original space approximates the closest vector (centroid $\mathcal{C}_i$) in the codebook.
\begin{equation}
\arg \min_{i \in k} \| \boldsymbol{v} - \mathcal{C}_{i} \|^2, \forall \boldsymbol{v} \in \mathbf{W'}
\label{eq:vq_eq}
\end{equation}
VQ indicates the nearest centroid $\mathcal{C}_{i}$ that minimizes the Euclidean distance between the input vector $\boldsymbol{v}$ in the lookup table. 
The optimization problem aims to find the index $i$ that results in the smallest distance between $\boldsymbol{v}$. 
Thus, each input vector is represented by the most similar centroids, thus minimizing total distortion.

Recent research has explored the use of VQ for model weight quantization \citep{diff-embed-pq, DKM, ABGD, PQ-metaseq}. These studies attempt to compress the embedding layer, the convolution layer, and the classification layer of neural networks using VQ. Figure~\ref{fig:vq_example} illustrates an example of applying VQ to compress model weights on a weight matrix.
For a weight matrix $\mathbf{W}$ with dimensions $M \times N$, we reshape $\mathbf{W}$ into vectors of length $\boldsymbol{v}$ as $\mathbf{W'}$ (step \ding{202}). 
The number of reshaped vectors should be $\frac{M \times N}{\boldsymbol{v}}$.
Next, we employ k-means or other clustering algorithms to build a codebook (step \ding{203}). 
The constructed codebook contains $k$ centroid vectors, each with $\boldsymbol{v}$ dimensions.
Applying the VQ algorithm directly often does not yield an acceptable accuracy. 
Typically, PTQ algorithms adjust the model index and centroid to enhance the accuracy of the quantized model (step \ding{204}).

During model inference, each operator in the model first dequantizes the original weight matrix from the lookup table (codebook) by index and centroid. 
Unlike scalar quantization, VQ keeps the index and centroid in quantized weight. 
The equivalent compression ratio of VQ can be formulated as: $\text{total original model bits} / (\text{codebook bits} + \text{index bits})$. 
The equivalent quantization bitwidth is as: $\text{original bit width}/\text{compression ratio}$.
For example, a $4096 \times 4096$ FP16 weight matrix with vectors of length $v=8$ and $256$ centroids, the compression ratio is $(16 \times 4096 \times 4096) / (8 \times 256 \times 16 + log_{2}(256) \times 4096 \times 4096 / 8) = 15.97$. 
The equivalent bitwidth is $1.002$ bit.

\subsection{Vector Quantization in LLMs}
While VQ has been applied to weight quantization, the following significant challenges persist when quantizing LLM. 
We summarize the benefits and weaknesses of recent research \citep{AQLM, QuIPsharp, GPTVQ} techniques in Table \ref{tab:intro_table}.

The number of parameters in LLMs is enormous, which requires quantizing the model using lightweight methods to avoid excessive resource consumption.
AQLM \citep{AQLM} utilizes gradient descent to train each layer of the VQ-quantized model and simultaneously trains across multiple layers using calibration data.
It achieves effective compression through additive quantization and joint optimization of the codebook, which can achieve high accuracy. However, due to AQLM's use of backpropagation for model training, significant GPU hours and memory are required to achieve better accuracy, especially when dealing with LLMs with massive parameters.

GPTVQ \citep{GPTVQ} utilizes the Second-Order Optimization method to implement PTQ. 
However, GPTVQ accumulates quantization errors within vector quantization, leading to an inevitable increase in quantization errors as the vector length increases. 
It prevents the use of longer vectors and consequently limits the compression ratio.

QuIP\# \citep{QuIPsharp} introduces an incoherence processing using the randomized Hadamard transform for the weight matrix before VQ. 
The processed weight matrix approximates a sub-Gaussian distribution, allowing for compression with a tiny codebook.
However, incoherence processing requires a significant amount of computation, despite QuIP\# being able to compress LLM to extremely low-bit with a low accuracy drop.
It requires significantly more computation for inference compared to the original LLM, resulting in low inference throughput.

\section{Vector Post-Training Quantization} \label{sec:vptq}
\vspace{-0.05in}

\begin{algorithm}[t!]
\small
\caption{VPTQ Algorithm}\label{alg:VPTQ}
\textbf{Input:} $\mathbf{W} \gets \mathbb{R}^{M \times N}$ \Comment{Input weight matrix} \\
\textbf{Input:} $\mathbf{H} \gets \mathbb{R}^{N \times N}$ \Comment{Hessian matrix} \\
\textbf{Output:} $\mathbf{\hat{W}} \gets \mathbb{R}^{M \times N}$ \Comment{Quantized weight matrix}

\begin{algorithmic}

\State \(\mathbf{E} \gets \mathbb{R}^{M \times N}\) \Comment{Initialize quantization errors}

\For{\(s = 0, B, 2B, \ldots\)} \Comment{Column blocks}
    \For{\(\text{n} = s, s+1, \ldots, s+B-1\)} \\\Comment{Quantize a single column $n$, fundamentally different from AQLM}
        \For{$m = 0, V, 2V, \ldots, M$} 
        \\\Comment{Parallel (Residual) Vector Quantization by function $Q(\boldsymbol{v})$ to vectors in the column $n$}
            \State $\mathbf{\hat{W}}_{m:m+V, n}$$\gets$ $Q_{V}(\mathbf{W}_{m:m+V,n})$
        \EndFor
        
        \State $\mathbf{E}_{:,n} \gets (\mathbf{W}_{:,n} - \mathbf{W'}_{:,n}) /(\mathbf{H}^{-1}_{n,n})$ 
        \\\Comment{Update quantization error}
        
        \State  $\mathbf{W}_{:, n:s+B} \gets \mathbf{W}_{:, n:s+B} - \mathbf{E}_{:,n} \mathbf{H}^{-1}_{n, n:s+B}$ \\
        \Comment{Merge quantization error to weights}
    \EndFor
    \State \(\mathbf{W}_{:,s+B:} \gets \mathbf{W}_{:,s+B:} - \mathbf{E}_{:, s:s+B} \mathbf{H}^{-1}_{s:s+B, s+B:}\)    \\ \Comment{Update all remaining weights}
\EndFor
\end{algorithmic}
\end{algorithm}

\subsection{VPTQ Algorithm} \label{sec:vptq_detail}
\vspace{-0.05in}

VPTQ leverages Second-Order Optimization and solves the optimization problem Eq.\ref{eq:optimization_problem} to achieve extreme low-bit quantization.
Assume that a weight matrix is $\mathbf{W} \in \mathbb{R}^{M \times N}$, and a Hessian matrix collected from the current layer is $\mathbf{H} \in \mathbb{R}^{N \times N}$. 
We denote the $q$-th column of the weight matrix as $\hat{\mathbf{W}}_{:,q}$. 
The quantized column $\hat{\mathbf{W}}_{:,q}$ can be represented as the transpose of concatenated centroid vectors
\begin{equation*}
    \hat{\mathbf{W}}_{:,q} = (\mathcal{C}_{0},\mathcal{C}_{1}, ... , \mathcal{C}_{M/v})^{T}.
\end{equation*}
When the weight matrix of the model is large, we can first split the weight matrix into multiple groups. Each group has its own independent codebook. This method allows us to flexibly divide the weight matrix into several submatrices ($\hat{\mathbf{W}}_{:,q:q+(M/\text{group num})}$) equal to the group number. 
For clarity, we describe only one group in the following algorithm description.

Unlike GPTVQ, we quantize each column of the matrix independently, which we refer to as \textbf{Channel-Independent Second-Order Optimization}. 
It greatly simplifies the complexity of VQ in Second-Order Optimization. 
GPTVQ, on the other hand, quantizes $\boldsymbol{v}$ columns of the matrix ($\hat{\mathbf{W}}_{M,v}$) at once, leading to larger errors and more complex transformations for problem optimization.

We use the Lagrange Method to transform the optimization problem \ref{eq:optimization_problem} into an unconstrained optimization problem. The Lagrangian function $L(\Delta \mathbf{W})$, and $\mathbf{\lambda}$ is the Lagrangian multiplier:
\begin{equation*}
    L(\Delta \mathbf{W}) = \Delta \mathbf{W}^T H (\mathbf{W}) \Delta \mathbf{W} + \mathbf{\lambda} \Delta \mathbf{W} \label{eq:lagrangian}
\end{equation*}
The dual function $g(\lambda)$ can be represented as:
\begin{equation*}
    g(\lambda) = -\mathbf{H}^{-1}_{qq} \lambda \lambda^T  - \lambda  (\hat{\mathbf{W}}_{:,q} - \mathbf{W}_{:,q}) 
\end{equation*}
Differentiating $g(\lambda)$ with respect to $\lambda$ and setting it to $0$, 
\begin{equation*}
    g'(\lambda) = -\mathbf{H}^{-1}_{qq} \lambda -  (\hat{\mathbf{W}}_{:,q} - \mathbf{W}_{:,q}) ^T = 0
\end{equation*}
we can find that when $\lambda^T = - \frac{(\hat{\mathbf{W}}_{:,q} - \mathbf{W}_{:,q})}{\mathbf{H}^{-1}_{qq}}$, the problem reaches an optimal solution.

By substituting $\lambda^T$ into the optimization problem, we find that to minimize the error introduced by quantization, we need to minimize the impact on the Lagrangian function. Therefore, we can transform the quantization problem into minimizing:
\begin{equation*}
    \Delta L (\Delta \mathbf{\hat{\mathbf{W}}}) = \frac{\sum \| \boldsymbol{v} - \mathcal{C} \|^2}{ 2 \mathbf{H}^{-1}_{qq}} \label{eq:lagrangian_obj}
\end{equation*}
We find that when quantizing a column vector each time, we only need to consider minimizing $\sum \| \boldsymbol{v} - \mathcal{C} \|^2$, which is to find the closest centroid in Euclidean Distance.
It precisely aligns with the optimization of VQ.
Moreover, since VPTQ quantizes the weight matrix column by column, $\mathbf{H}^{-1}_{qq}$ is constant when quantizing each column, so we do not need to consider Hessian when finding the centroid.

After quantizing a column of the weight matrix, we need to update the current quantization error to the unquantized part through:
\begin{equation*}
    \begin{aligned}
    \Delta \mathbf{W} = \frac{(\mathbf{\hat{W}}_{:,q} - \mathbf{W}_{:,q})}{\mathbf{H}^{-1}_{qq}} \mathbf{H}_{q,:}
    \end{aligned}
\end{equation*}
It will transform current quantization errors to the following unquantized columns. 
Since GPTVQ quantizes $\boldsymbol{v}$ columns at the same time, quantization error can only spread to other unquantized columns when all $\boldsymbol{v}$ columns have been quantized.
It will lead to more errors accumulating in the quantization, resulting in a decrease in model accuracy. 
We can have similar conclusions from Table \ref{tab:llama2_2bit_avg}. 
Algorithm \ref{alg:VPTQ} provides a detailed description of the steps to solve the optimization problem and quantize the weights according to the above analysis.

\textbf{Distinguish VPTQ from GPTQ and GPTVQ:}
Compared with GPTQ, VPTQ employs vector representations in the quantization, which choose the vector closest to the original matrix to represent the original data. 
As VQ can use a larger codebook to store the quantized data, it covers a wider range of numerical distributions compared to the scalar quantization of GPTQ, thereby achieving better accuracy. 
Table \ref{tab:llama2_2bit_avg} reveals that VPTQ significantly outperforms GPTQ under extremely low bit quantization.

Moreover, since GPTVQ quantizes multiple columns simultaneously, the propagation of quantization errors to unquantized columns is more challenging.
Furthermore, the quantization errors in GPTVQ accumulate as the vector length increases, hindering GPTVQ from using longer vector lengths for weight compression (limited to only 1-4 bits). 
It significantly reduces the compression ratio of VQ. 
On the other hand, VPTQ is capable of compressing weights using longer vectors (> $8$ bits) and representing data with a larger codebook. Table \ref{tab:llama2_2bit_avg} shows the better accuracy achieved by VPTQ than GPTVQ.

\subsection{Optimization in VPTQ} \label{sec:vptq_optimization}

\subsubsection{Hessian-Weighted Centroid Initialization} \label{sec:hessian_weight_centroid}

VTPQ algorithm requires the initialization of centroids in the codebooks prior to quantization. 
Properly initializing centroids can reduce quantization errors and improve model accuracy. 
A straightforward method is to perform K-means clustering on the weight matrix as centroids (Eq.\ref{eq:vq_eq}). 
However, it does not consider the optimization object in Eq.\ref{eq:optimization_problem}, leading to a significant accuracy drop \citep{GPTVQ, AQLM}.

We can transform the optimization object by leveraging the cyclic property of matrix traces and the Hadamard product. We refine the optimization objective as:
\begin{equation*}
    \begin{aligned}
        &   \Delta \mathbf{W}^T \Delta \mathbf{W} \odot \mathbf{H} 
            = \sum_{i=0}^{n-1} h_{i,i} \|\Delta \mathbf{W}_{:,i} \|^2 \\ 
        &   + \sum_{i=0}^{n-1} \sum_{j=0,j \neq i }^{n-1} h_{i,j} (\Delta \mathbf{W}_{:,i} \Delta \mathbf{W}_{:,j})
    \end{aligned}
    \label{eq:proxy_error}
\end{equation*}
Due to the Hessian matrix being predominantly diagonal \cite{hessia_diag}, it guides us to split the proxy error into two terms. The first term represents the dominant diagonal elements of the initial error matrix, which significantly impact the quantization error. 
The second term is the interaction of a single value in weight quantization with others. 

Because the Hessian matrix is predominantly diagonal, we can prioritize optimizing the first term through centroid initialization. We can view the first term as a Weighted K-means Clustering problem \citep{weighted_kmeans, weighted_kmeans2, weighted_kmeans3}. 
Since this problem is well-studied, we can directly solve it to achieve efficient and accurate centroid initialization.

\subsubsection{Residual Vector Quantization}
We enable Residual Vector Quantization (RVQ) \citep{residual_vq, residual_vp2} in VPTQ. RVQ improves vector quantization (VQ) by breaking down the compression of a weight matrix into two (or more) stages. 
Each stage further compresses the residual error $v_{\text{res}} = v - Q(v)$ from the previous quantization stage:
\begin{equation*}
    Q(\boldsymbol{v}_{\text{res}}) = \arg \min_{i} \| (\boldsymbol{v} - Q(\boldsymbol{v})) - \mathcal{C}^{\text{res}}_{i} \|^2
\end{equation*}

Unlike GPTVQ, VPTQ enables RVQ, which quantizes VQ quantization error using a separate lookup table for better representation and quantization.
By partitioning the encoding into multiple stages and reducing quantization error, RVQ not only achieves superior compression efficiency but also ensures a balance between quantization error, the size of lookup tables, and the memory requirements for indices. 
During the decoding phase, VPTQ simply reads the centroids from these multiple lookup tables and combines them to reconstruct the original weight matrix.

\subsubsection{Outlier Elimination}
Recent studies on quantization in LLM have consistently observed a significant presence of outliers in activation \citep{SmoothQuant, awq, owq}.  
Outliers, while small portions (\textasciitilde$1\%$ of the matrix), heavily affect the quantization error and simulate model accuracy.
Outliers typically result in large values in the diagonal elements of the Hessian matrix. 
During centroid initialization in Sec.\ref{sec:hessian_weight_centroid}, VPTQ already considers these Hessian diagonals as weights in K-means, allowing VPTQ to better quantize the error introduced by outliers.
\begin{equation*}
    Q(\boldsymbol{v}_{\text{outlier}}) = \arg \min_{i} \| \boldsymbol{v}_{\text{outlier}} - \mathcal{C}^{\text{outlier}}_{i} \|^2
\end{equation*}
Furthermore, VPTQ flexibly partitions the weight matrix and uses a separate outlier lookup table to quantify matrix tiles most affected by outliers.
It allows us to effectively trade off model accuracy and quantization overhead.

\section{End to end Quantization Algorithm}\label{sec:e2e}
\vspace{-0.05in}

\begin{algorithm}[t!]
\caption{End to End Quantization Algorithm}\label{alg:cap}
\small
\begin{algorithmic}
\Require original model, vector length $v$, centroid number $k$, hessian matrices $\mathbf{H}$
\Ensure quantized model
\For{each layer $l$} \Comment{Fully parallelized each layer on GPUs}
    \For{each Linear operator}
        \If{outlier is enabled}
            \State Initialize outlier centroids $\mathcal{C}_{\text{outlier}}$ 
            \State $\mathbf{W}'_{\text{outlier}} \gets \text{VPTQ}(\mathbf{W}_{\text{outlier}}, \mathcal{C}_{\text{outlier}})$
        \EndIf
        \State Initialize centroids $\mathcal{C}$ 
        \State $w' \gets \text{VPTQ}(\mathbf{W}, \mathcal{C})$
        \If{residual is enabled}
            \State Initialize residual centroids $\mathcal{C}_{\text{res}}$ 
            \State $\mathbf{W}'' \gets \text{VPTQ}(\mathbf{W} - \mathbf{W}', \mathcal{C}_{\text{res}})$
        \EndIf
    \EndFor
    \If{finetune layer is enabled}
        \State Finetune layer $l$
    \EndIf
\EndFor
\end{algorithmic}
\end{algorithm}

In this section, we will detail the end-to-end model quantization algorithm (Algorithm \ref{alg:cap}). 
The algorithm takes the original model, vector length \(v\), centroid number \(k\), and Hessian matrices \(\mathbf{H}\) as inputs. 
It starts by iterating over each layer \(l\) of the model. As each layer's quantization only relates to the current layer and the Hessian matrix, we can fully parallelize the quantization of each layer on GPUs.

In each layer, we first quantize the weight of each Linear Operator (matrix multiplication of input and weight). If we enable the outlier option, the algorithm first selects outlier columns following Section \ref{sec:vptq_optimization} and initializes the outlier centroids \(\mathcal{C}_{\text{outlier}}\). 
Then, VPTQ is applied to the outlier weights \(\mathbf{W}_{\text{outlier}}\) using the outlier centroids, generating the quantized weights \(\mathbf{W}'_{\text{outlier}}\). 
Next, the algorithm initializes the centroids \(\mathcal{C}\) for the remaining columns and applies VPTQ to the weights \(\mathbf{W}\) using these centroids to produce the quantized weights \(\mathbf{W}'\). 
Lastly, if residual quantization is enabled, the algorithm initializes the residual centroids \(\mathcal{C}_{\text{res}}\). 
It applies VPTQ to the residual error between the original weights and the quantized weights (\(\mathbf{W} - \mathbf{W}'\)), using the residual centroids.
The quantized weight is updated as \(\mathbf{W}''\).

After processing all the operators, the algorithm will fine-tune the layer \(l\) if we enable layer fine-tuning. 
The loss function is the Mean Squared Error (MSE) between the original and quantized computations. 
In layer-wise fine-tuning, we only update the normalization operator (e.g. RMSNorm) and centroid. 
These parameters only comprise a small fraction of the entire layer, and we can complete the fine-tuning quickly with limited memory. After each layer completes quantization and fine-tuning, we can further fine-tune the entire model as other PTQ methods used \citep{QuIPsharp, QuIP, AQLM}.
Once the algorithm processes all layers, it outputs the quantized model. 
The end-to-end VPTQ algorithm quantizes all the weights in every Linear Operator in the model into an index and a codebook (\(\mathcal{C}\)). 
During model inference, we only need to dequantize the weight matrix by reading centroids from the codebook according to the index before executing the operator.

\section{Experiments and Evaluations}
\vspace{-0.05in}
\begin{table*}[hbt!]
\centering
\caption{LLaMA-2 2bit Quantization Results. The "N/A" in the table stands for "not available," with further explanation provided in the Appendix \ref{appendix:na_explain}.
\footnotesize{\footnotemark[1] We use the naive Torch and Triton kernels for inference performance evaluation, without optimizations like CUDA graphs, FlashAttention, or Torch compile. The inference performance for QuIP\# and AQLM do not represent their performance with all optimizations enabled. QuIP\# and AQLM can achieve high performance when all optimizations are enabled.}} 
\label{tab:llama2_2bit_avg}

\begin{subtable}{\textwidth}
\centering
\caption{7B results}
\resizebox{0.65\textwidth}{!}{%
\begin{tabular}{c|ccccccc}
\hline
Method & bit   & W2↓           & C4↓           & AvgQA↑        & tok/s↑        & mem(GB)↓      & cost(h)↓ \\ \hline
FP16   & 16    & 5.12          & 6.63          & 62.2          & 38.32         & 27.22         & N/A      \\
GPTQ   & 2.125 & 50.75         & 36.76         & 39.16         & 19.59         & 4.42          & 0.2      \\
GPTVQ  & 2.25  & 6.71          & 9.9           & 56.14         & N/A           & N/A           & 1.5      \\
DB-LLM & 2.01  & 7.23          & 9.62          & 55.1          & N/A           & N/A           & N/A      \\
QuIP\#\footnotemark[1] & 2     & 6.19          & 8.16          & \textbf{58.2}          & 4.4           & 2.25 & N/A      \\
AQLM \footnotemark[1]   & 2.02  & 6.64          & 8.56          & 56.5 & 19.4          & \textbf{2.16}          & N/A      \\
AQLM \footnotemark[1]   & 2.29  & 6.29          & 8.11          & 58.6          & 19.6          & 2.4           & 11.07    \\ \hline
VPTQ   & 2.02  & \textbf{6.13} & \textbf{8.07} & \textbf{58.2}          & \textbf{39.9} & 2.28          & 2        \\
       & 2.26  & \textbf{5.95} & \textbf{7.87} & \textbf{59.4} & \textbf{35.7} & 2.48          & 2.2      \\ \hline
\end{tabular}

}
\end{subtable}

\vspace{1em} %

\begin{subtable}{\textwidth}
\centering
\caption{13B results}
\resizebox{0.65\textwidth}{!}{%
\begin{tabular}{c|ccccccc}
\hline
Method                & bit   & W2↓           & C4↓           & AvgQA↑        & tok/s↑        & mem(GB)↓      & cost(h)↓ \\ \hline
FP16                  & 16    & 4.57          & 6.05          & 65.4          & 30.03         & 63.63         & N/A      \\
GPTQ                  & 2.125 & 43.84         & 23.07         & 43.72         & 11.56         & 7.92          & 0.3      \\
GPTVQ                 & 2.25  & 5.72          & 8.43          & 61.56         & N/A           & N/A           & 3.7      \\
DB-LLM                & 2.01  & 6.19          & 8.38          & 59.4          & N/A           & N/A           & N/A      \\
QuIP\#\footnotemark[1]                & 2     & 5.35          & 7.2           & 62.0          & 3.5           & \textbf{3.94} & N/A      \\
AQLM \footnotemark[1]                 & 1.97  & 5.65          & 7.51          & 60.6          & N/A           & N/A           & N/A      \\
AQLM \footnotemark[1]                  & 2.18  & 5.41          & 7.2           & 61.6          & 16.5          & 4.14          & 22.7     \\ \hline
VPTQ                  & 2.02  & \textbf{5.32} & \textbf{7.15} & \textbf{62.4} & \textbf{26.9} & 4.03          & 3.2      \\
\multicolumn{1}{l|}{} & 2.18  & \textbf{5.28} & \textbf{7.04} & \textbf{63.1} & \textbf{18.5} & 4.31          & 4        \\ \hline
\end{tabular}
}
\end{subtable}

\vspace{1em} %

\begin{subtable}{\textwidth}
\centering
\caption{70B results}
\resizebox{0.65\textwidth}{!}{%
\begin{tabular}{c|ccccccc}
\hline
Method & bit   & W2↓           & C4↓           & AvgQA↑        & tok/s↑       & mem(GB)↓       & cost(h)↓ \\ \hline
FP16   & 16    & 3.12          & 4.97          & 70.2          & multi-gpu    & N/A            & N/A      \\
GPTQ   & 2.125 & NaN           & NaN           & 59.18         & 2.38         & 37.63          & 2.83     \\
GPTVQ  & 2.25  & 4.25          & 6.9           & 68.5          & N/A          & N/A            & 12       \\
DB-LLM & 2.01  & 4.64          & 6.77          & 65.8          & N/A          & N/A            & N/A      \\
QuIP\#\footnotemark[1] & 2     & \textbf{3.91} & \textbf{5.71} & \textbf{69.0} & 1.9          & \textbf{18.36} & 25      \\
AQLM \footnotemark[1]   & 2.07  & 3.94          & 5.72          & 68.8          & 6.9          & 18.81          & 183      \\ \hline
VPTQ   & 2.07  & 3.93          & 5.72          & 68.6          & \textbf{9.7} & 19.54          & 19       \\
VPTQ   & 2.11  & 3.92          & \textbf{5.71} & 68.7          & \textbf{9.7} & 20.01          & 19       \\ \hline
\end{tabular}
}
\end{subtable}
\end{table*}
\begin{table*}[hbt!]
\centering
\caption{LLaMA-3 and Mistra-7b 2,3,4-bit Quantization Results. The table shows LLaMA-3 Wikitext2 perplexity (context length 2048) and average zero-shot QA Accuracy, Mistral-7B Wikitext2, C4 perplexity (context length 8192) and average zero-shot QA accuracy. Detailed score for each task see Table \ref{tab:llama3-234bit} and Table \ref{tab:mistral-7b-234bit}.}
\vspace{-0.1in}
\label{tab:llama3-mistral-avg}
\resizebox{0.9\textwidth}{!}{%
\begin{tabular}{c|ccc|ccc|c|cccc}
\hline
\multirow{4}{*}{} & \multicolumn{3}{c|}{\multirow{2}{*}{LLaMA-3 8B}}                      & \multicolumn{3}{c|}{\multirow{2}{*}{LLaMA-3 70B}}                     & \multirow{2}{*}{} & \multicolumn{4}{c}{\multirow{2}{*}{Mistral 7B}}                                              \\
                  & \multicolumn{3}{c|}{}                                                 & \multicolumn{3}{c|}{}                                                 &                   & \multicolumn{4}{c}{}                                                                         \\ \cline{2-7} \cline{9-12} 
                  & \multirow{2}{*}{bit} & \multirow{2}{*}{W2↓} & \multirow{2}{*}{AvgQA↑} & \multirow{2}{*}{bit} & \multirow{2}{*}{W2↓} & \multirow{2}{*}{AvgQA↑} & \multirow{2}{*}{} & \multirow{2}{*}{bit} & \multirow{2}{*}{W2↓} & \multirow{2}{*}{C4↓} & \multirow{2}{*}{AvgQA↑} \\
                  &                      &                      &                         &                      &                      &                         &                   &                      &                      &                      &                         \\ \hline
FP16              & 16                   & 6.14                 & 68.6                    & 16                   & 2.9                  & 75.3                    & FP16              & 16.0                 & 4.77                 & 5.71                 & 68.6                    \\ \hline
QuIP              & 4                    & 6.5                  & 67.1                    & 4                    & 3.4                  & 74.5                    & QuIP\#            & 4.01                 & 4.85                 & 5.79                 & \textbf{68.7}           \\
GPTQ              & 4                    & 6.5                  & 67.3                    & 4                    & 3.3                  & \textbf{74.9}           & AQLM              & 4.02                 & 4.85                 & 5.79                 & 68.0                    \\
VPTQ              & 4.03                 & \textbf{6.42}        & \textbf{68.1}           & 4.05                 & \textbf{3.15}        & 74.7                    & GPTQ              & 4.125                & 4.83                 & 5.74                 & 68.4                    \\ \cline{1-7}
QuIP              & 3                    & 7.5                  & 63.7                    & 3                    & 4.7                  & 72.6                    & VPTQ              & 4.03                 & \textbf{4.81}        & \textbf{5.72}        & 68.2                    \\ \cline{8-12} 
GPTQ              & 3                    & 8.2                  & 61.7                    & 3                    & 5.2                  & 70.6                    & AQLM              & 3.0                  & 5.07                 & 5.97                 & \textbf{67.3}           \\
VPTQ              & 3.03                 & \textbf{6.97}        & \textbf{66.7}           & 3.01                 & \textbf{3.81}        & \textbf{73.7}           & VPTQ              & 3.03                 & \textbf{4.96}        & \textbf{5.84}        & \textbf{67.3}           \\ \hline
QuIP              & 2                    & 85.1                 & 36.8                    & 2                    & 13                   & 48.7                    & QuIP\#            & 2.01                 & 6.02                 & 6.84                 & 62.2                    \\
DB-LLM            & 2                    & 13.6                 & 51.7                    & N/A                  & N/A                  & N/A                     & AQLM              & 2.01                 & 6.32                 & 6.93                 & 62.2                    \\
GPTQ              & 2                    & 2.10E+02             & 36.2                    & 2                    & 11.9                 & 45.4                    & GPTQ              & 2.125                & 1535                 & 164                  & 44.5                    \\
VPTQ              & 2.08                 & \textbf{9.29}        & \textbf{60.2}           & 2.02                 & \textbf{5.6}         & \textbf{70.9}           & GPTVQ             & 2.25                 & 8.99                 & 18.6                 & 57.7                    \\
VPTQ              & 2.24                 & \textbf{9.19}        & \textbf{62.7}           & 2.07                 & \textbf{5.66}        & \textbf{70.7}           & VPTQ              & 2.04                 & \textbf{5.64}        & \textbf{6.43}        & \textbf{63.2}           \\ \hline
\end{tabular}
}
\vspace{-0.2in}
\end{table*}

\subsection{Settings}
\vspace{-0.05in}
 \label{sec:exp_settings}
\quad\textbf{Algorithm Baseline} We focus on weight-only quantization. The detailed quantization parameters (such as vector length and codebook numbers) and fine-tuning parameters of our VPTQ are shown in Appendix \ref{appendix:analysis} . Following \cite{gptq}, our calibration data consists of 128 random segments of the C4 dataset \citep{c4}. 

\textbf{Models and Datasets} We benchmark accuracy on LLaMA-2 \citep{llama2}, LLaMA-3 families \citep{llama3}, and Mistral \cite{mistral7b}. 
Following previous work \citep{gptq}, we report perplexity on language modeling tasks (WikiText-2 \citep{wikitext2}, C4 \citep{c4}).  We also employ lm-eval-harness \citep{lm_eval} to perform zero-shot evaluations on common sense QA benchmarks (PIQA \citep{piqa}, HellaSwag \citep{hellaswag}, WinoGrande \citep{winogrande}, ARC \citep{arc}). Detailed configuration is in Appendix \ref{appdx}.

\textbf{Baselines} For LLaMA-2 and Mistral models, we compare VPTQ against GPTQ, GPTVQ, DB-LLM, QuIP\#, and AQLM.  
To account for the different overheads resulting from varying codebook constructions, we provide results with comparable bit widths to facilitate a fair comparison.
For LLaMA-3 models, we use the results of \cite{llama3quantization}. 
However, due to alignment issues with the C4 dataset, we only show results for WikiText and QA tasks. Because LLaMA-3 models are new and running quantization ourselves is costly, we do not have results for QuIP\# and AQLM.

\subsection{Accuracy Evaluation}
\vspace{-0.05in}
\textbf{Results on LLaMA-2 model:} 
We compare VPTQ with QuIP\#, AQLM, GPTVQ, DB-LLM, and GPTQ on the LLaMA-2 model. 
First, we discuss the results of 2-bit quantization. As shown in Table \ref{tab:llama2_2bit_avg}, GPTQ, as a scalar quantization method, performs poorly with unusable accuracy. 
While DB-LLM and GPTVQ perform better, they still experience significant performance drops, with WikiText-2 perplexity increasing by 2.  The significant accuracy drop in GPTVQ, despite being a vector quantization algorithm, is due to two factors: the use of shorter vector lengths, which introduces higher quantization loss, and the choice to update weights every $v$ columns, which leads to cumulative errors. Therefore, we primarily focus on comparing VPTQ with the state-of-the-art QuIP\# and AQLM which both choose longer vector lengths. 

Table \ref{tab:llama2_2bit_avg} includes the average scores for the five QA tasks mentioned in Section \ref{sec:exp_settings}.
VPTQ outperforms QuIP\# and AQLM on 7B and 13B models. 
For the 7B model, VPTQ achieves a further reduction in WikiText-2 perplexity by 0.5 and 0.3 compared to the previous best results %
at 2-2.02 bits and 2.26-2.29 bits, respectively. In QA tasks, the VPTQ 2.26-bit model surpasses the AQLM 2.29-bit model with an average accuracy increase of $1\%$. For the 13B model, the VPTQ 2.02-bit model shows a slight improvement over QuIP\#, and the 2.18-bit model outperforms AQLM in QA accuracy by $1.5\%$.
On the LLaMA-2-70B model, we achieve similar perplexity ($<0.02$) and comparable QA results($<0.4\%$). The results for 3- and 4-bit quantization shown in Table \ref{tab:llama2-34bit} are without end-to-end fine-tuning but are also comparable to AQLM and QuIP\# which include end-to-end fine-tuning. The ablation study of quantization parameters is in Appendix \ref{sec:ablation_study}. 

\textbf{Results on LLaMA-3 and Mistral model:} 
Table \ref{tab:llama3-mistral-avg} presents VPTQ results on the LLaMA-3 model and Mistral-7b model. 
In all 2-, 3-, and 4-bit quantizations of LLaMA-3 models, we significantly outperform GPTQ, DB-LLM, and QuIP, whose accuracy drops to unusable levels. VPTQ ensures an accuracy drop of $<8\%$ for the 8B model and $<5\%$ for the 70B model. 
On the Mistral-7B model, our 2-bit performance surpasses both QuIP\# and AQLM by $0.8\%$ in QA accuracy. In 3-bit quantization, our perplexity is lower. 
At 4-bit, results are comparable overall. More detailed results are in Table \ref{tab:mistral-7b-234bit}. As bit width increases, the advantage of vector quantization diminishes, with GPTQ showing a similar WikiText-2 perplexity at 4-bit.

\textbf{Inference throughput and quantization cost:}
In Table \ref{tab:llama2_2bit_avg}, the `tok/s' column indicates the number of tokens generated per second during the decode phase of inference. VPTQ achieves a $2$-$9\times$ speedup compared to QuIP\# because QuIP\# uses Hadamard Transform during decoding, which introduces $O(n^2)$ multiplications and additions, significantly slowing the inference throughput. Compared to AQLM, VPTQ uses a smaller codebook, resulting in a lower decoding overhead. Therefore, our inference throughput for the 7B and 13B models is $1.6$-$1.8\times$ faster than AQLM. As the model size increases, our codebook size becomes comparable to theirs, leading to similar inference throughputs for the 70B model. The 'mem(GB)' column represents the GPU memory usage at runtime.
The `cost(h)' column represents the hours required for model quantization on $4\times$ 80GB A100 GPUs. We achieves comparable or even better results than AQLM in only $10.4$-$18.6\%$ of quantization algorithm execution time.

\section{Conclusion}
\vspace{-0.05in}
In this paper, we propose Vector Post-Training Quantization (VPTQ), a novel approach to achieving extremely low-bit quantization of LLMs by Vector Quantization. 
Through the application of Second-Order Optimization, we have formulated the LLM Vector Quantization problem and directed the design of our quantization algorithm. By further refining the weights via Channel-Independent Second-Order Optimization, we have enabled a more granular VQ. 

VPTQ also includes a brief and effective codebook initialization algorithm, which is achieved by decomposing the optimization problem. We have extended VPTQ to support residual and outlier quantization, which not only improves model accuracy but also further compresses the model size. 

Our experimental results demonstrate the effectiveness and efficiency of VPTQ. 
The perplexity of quantized model is reduced by $0.01$-$0.34$ on LLaMA-2, $0.38$-$0.68$ on Mistral-7B, $4.41$-$7.34$ on LLaMA-3 over SOTA at 2-bit, with an average accuracy improvement of  $0.79$-$1.5\%$ on LLaMA-2, $1\%$ on Mistral-7B, $11$-$22\%$ on LLaMA-3  on QA tasks. 
Furthermore, we achieved these results only using $10.4$-$18.6\%$ of the execution time of the quantization algorithm, leading to a $1.6$-$1.8\times$ increase in inference throughput compared to SOTA. 
These results underscore the potential of VPTQ as an efficient and powerful solution for the deployment and inference of LLMs, particularly in resource-constrained settings.

\section{Limitations}
Related research on PTQ \citep{AQLM, QuIPsharp, GPTVQ} have adopted end-to-end model fine-tuning after the PTQ phase. Compared to other related works, VPTQ can better quantize the model in the PTQ, and it simplifies and reduces the cost and overhead of model fine-tuning.

Due to GPU resource constraints, we cannot fine-tune larger models (70B) for longer iterations and more tokens. 
It limits our experimental results, which can only achieve similar results to baselines in 70B models. It restricts the demonstration of VPTQ's advantages and potential on large models in this paper. 
We will strive for more GPU resources to fine-tune the VPTQ model for longer periods and with more tokens in the future, allowing for a fair comparison.

Additionally, since LLaMA-3 models are the latest released models, there is a lack of baselines from related works. 
It is difficult for us to fully demonstrate our performance improvements. 
We will continue to add more baselines in the future to highlight the advantages of VPTQ.

In this paper, we only use AI tools for grammar checking and code completion.

\vspace{-0.1in}
\section*{Acknowledgement}
\vspace{-0.1in}
We thank James Hensman for his crucial insights into the error analysis related to Vector Quantization (VQ), and his comments on LLMs evaluation are invaluable to this research. We also thank QuIP\# and AQLM for inspiring our paper and the authors for their guidance on implementation.

\bibliography{references}
\bibliographystyle{acl_natbib}

\appendix
\section{Appendix: All Experiments Results}
\label{appdx}

\subsection{Supplementary Explanation for Main Results Table 2}
\label{appendix:na_explain}
Table \ref{tab:llama2_2bit_avg} shows our main results. Here we provide an explanation for the 'N/A' entries relative to other works.

\textbf{DB-LLM} Since they did not open source their code, we use the AvgQA results from their paper. However, this number does not align with our FP16 results.

\textbf{GPTQ} We reproduce the 2-bit results using the official GPTQ repository. As GPTQ quantizes each layer in sequential order, the 'cost(h)' represents the time taken to quantize on a single A100 GPU.

\textbf{GPTVQ} They do not release their 2-bit quantized model. We reproduce Llama-2, LLama-3 7B and 13B, Mistral 7b 2-bit results using their released GPTVQ code, which only supports single-GPU execution. Therefore, the quantization cost reflects the execution time for quantization on a single A100 GPU. Due to the lack of specific logic for loading their quantizers in the released code, we were unable to measure the throughput and runtime memory.

\textbf{AQLM} Their 1.97-bit LLaMA-2 13b model has not been open-sourced, so we are unable to measure its inference throughput and runtime memory.

\textbf{QuIP\#} Due to recent changes in the libraries they rely on, the quantization cost is not measured. The quantization time for the 70B model is estimated based on their original paper.

\subsection{All Experimental Results}
In this section, we present all our experimental results, including the perplexity of the quantized model on different context lengths in two datasets, Wikitext2 and C4, and the accuracy on five Commonsense QA tasks (abbreviated as AE for Arc\_easy, AC for Arc\_challenge, HE for Hellaswag, QA for PIQA, and WI for Winogrande). Table \ref{tab:llama2-2bit-all} displays all results of LLaMA-2 at 2-bit quantization. Table \ref{tab:llama2-34bit} presents results of LLaMA-2 at 3 and 4 bits quantization. Table \ref{tab:llama3-234bit} displays all results of Llama3 at 2, 3, and 4-bit quantization. Table \ref{tab:mistral-7b-234bit} shows all results of Mistral 7b at 2, 3, and 4-bit quantization.

\begin{table*}[tbh!]
\caption{LLaMA-2 2bit Quantization Results, \footnotesize{\footnotemark[1] We use the naive Torch and Triton kernels for inference performance evaluation, without optimizations like CUDA graphs, FlashAttention, or Torch compile. The inference performance for QuIP\# and AQLM do not represent their performance with all optimizations enabled. QuIP\# and AQLM can achieve high performance when all optimizations are enabled.}}
\label{tab:llama2-2bit-all}
\centering
\resizebox{0.85\textwidth}{!}
{
\begin{tabular}{c|cccccccc|ccc}
\hline
7B     & bit   & W2            & C4            & AC             & AE             & HE             & QA             & WI             & tok/s         & mem(GB)        & cost(h) \\ \hline
FP16   & 16    & 5.12          & 6.63          & 39.93          & 69.28          & 56.69          & 78.35          & 66.93          & 38.32         & 27.22          & N/A     \\
GPTQ   & 2.125 & 50.75         & 36.76         & 20.9           & 34.9           & 30.5           & 57.2           & 52.3           & 19.59         & 4.42           & 0.2     \\
GPTVQ  & 2.25  & 6.71          & 9.9           & 31.2           & \textbf{66.3}  & 46.4           & 72.4           & 64.4           & N/A           & N/A            & 1.5     \\
DB-LLM & 2.01  & 7.23          & 9.62          & 33.53          & 45.2           & 61.98          & 73.18          & 61.72          & N/A           & N/A            & N/A     \\
QuIP\#\footnotemark[1] & 2     & 6.19          & 8.16          & 34.6           & 64.6           & 51.91          & 75.1           & \textbf{64.9}  & 4.4           & 2.25           & N/A      \\
AQLM\footnotemark[1]   & 2.02  & 6.64          & 8.56          & 33.28          & 61.87          & 49.49          & 73.56          & 64.17          & 19.4          & \textbf{2.16}  & N/A      \\
       & 2.29  & 6.29          & 8.11          & 34.9           & 66.5           & 50.88          & 74.92          & 65.67          & 19.6          & 2.4            & 11.07   \\
VPTQ   & 2.02  & \textbf{6.13} & \textbf{8.07} & \textbf{35.24} & 63.8           & 52.08          & \textbf{75.19} & 64.33          & \textbf{39.9} & 2.28           & 2       \\
       & 2.26  & \textbf{5.95 }         & \textbf{7.87}          & \textbf{36.43} & 64.9           & \textbf{52.87} & \textbf{76.17} & \textbf{66.46} & \textbf{35.7} & 2.48           & 2.2     \\ \hline
13b    & bit   & W2            & C4            & AC             & AE             & HE             & QA             & WI             & tok/s         & mem(GB)        & cost(h) \\ \hline
FP16   & 16    & 4.57          & 6.05          & 45.56          & 73.23          & 59.71          & 78.73          & 69.69          & 30.03         & 63.63          & N/A     \\
GPTQ   & 2.125 & 43.84         & 23.07         & 23.3           & 43.3           & 36             & 61.3           & 54.7           & 11.56         & 7.92           & 0.3     \\
GPTVQ  & 2.25  & 5.72          & 8.43          & 38.7           & \textbf{73.6}  & 51.6           & 75.4           & \textbf{68.5}  & N/A           & N/A            & 3.7     \\
DB-LLM & 2.01  & 6.19          & 8.38          & 38.14          & 51.64          & 68.04          & 75.14          & 64.09          & N/A           & N/A            & N/A     \\
QuIP\#\footnotemark[1] & 2     & 5.35          & 7.2           & 39.5           & 69.3           & 56.01          & \textbf{77.3}  & 67.7           & 3.5           & \textbf{3.94}           & N/A     \\
AQLM\footnotemark[1]   & 1.97  & 5.65          & 7.51          & 37.8           & 69.78          & 53.74          & 76.22          & 65.43          & N/A           & N/A            & N/A     \\
       & 2.18  & 5.41          & 7.2           & 39.42          & 69.15          & 54.68          & 76.22          & 68.43          & 16.5          & 4.14           & 22.7    \\
VPTQ   & 2.02  & \textbf{5.32} & \textbf{7.15} & \textbf{40.02} & 71.55          & \textbf{56.18} & 77.26          & 66.85          & \textbf{26.9} & 4.03           & 3.2     \\
       & 2.18  & \textbf{5.28} & \textbf{7.04} & \textbf{40.96} & 71.8           & \textbf{56.89} & \textbf{77.48} & 68.43          & \textbf{18.5} & 4.31           & 4       \\ \hline
70b    & bit   & W2            & C4            & AC             & AE             & HE             & QA             & WI             & tok/s         & mem(GB)        & cost(h)  \\ \hline
FP16   & 16    & 3.12          & 4.97          & 51.11          & 77.74          & 63.97          & 81.12          & 77.11          & multi-gpu     & N/A            & N/A     \\
GPTQ   & 2.125 & NaN           & NaN           & 35.8           & 67             & 51.8           & 74.6           & 66.7           & 2.38          & 37.63          & 2.83    \\
GPTVQ  & 2.25  & 4.25          & 6.9           & 49.4           & \textbf{80.47} & 58.26          & 79.4           & 75.2           & N/A           & N/A            & 12      \\
DB-LLM & 2.01  & 4.64          & 6.77          & 44.45          & 55.93          & 76.16          & 79.27          & 73.32          & N/A           & N/A            & N/A     \\
QuIP\#\footnotemark[1] & 2     & \textbf{3.91} & \textbf{5.71} & \textbf{48.7}  & 77.3           & \textbf{62.49} & 80.3           & 75.9           & 1.9           & \textbf{18.36} & 25     \\
AQLM\footnotemark[1]   & 2.07  & 3.94          & 5.72          & 47.93          & 77.68          & 61.79          & \textbf{80.43} & \textbf{75.93} & 6.9           & 18.81          & 183     \\
VPTQ   & 2.07  & 3.93          & 5.72          & 47.7           & 77.1           & \textbf{62.98} & 80.3           & 74.98          & \textbf{9.7}  & 19.54          & 19      \\
       & 2.11  & 3.92          & \textbf{5.71} & 48.29          & 77.77          & 62.51          & 79.82          & 75.14          & \textbf{9.7}  & 20.01          & 19      \\ \hline
\end{tabular}
}
\end{table*}
\begin{table*}[hbt!]
\centering
\captionsetup{justification=centering}
\caption{LLaMA-2 3, 4-bit Quantization Results. The table shows Witext2,\\ C4 perplexity (context length 2048 and 4096) and
zeroshot QA Accuracy.}
\label{tab:llama2-34bit}
\resizebox{0.8\textwidth}{!}{%
\begin{tabular}{c|c|cccc|ccccc}
\hline
                     &                      &                         &                         &                         &                         &                     &                     &                     &                     &                     \\
\multirow{2}{*}{7B}  & \multirow{2}{*}{bit} & \multirow{2}{*}{W2(2k)} & \multirow{2}{*}{C4(2k)} & \multirow{2}{*}{W2(4k)} & \multirow{2}{*}{C4(4k)} & \multirow{2}{*}{AC} & \multirow{2}{*}{AE} & \multirow{2}{*}{HE} & \multirow{2}{*}{QA} & \multirow{2}{*}{WI} \\
                     &                      &                         &                         &                         &                         &                     &                     &                     &                     &                     \\ \hline
GPTQ                 & 4                    & ---                     & ---                     & 5.49                    & 7.2                     & 36.8                & 66.2                & 55.4                & 76.6                & 68.2                \\
GPTVQ                & 4.125                & 5.68                    & 7.25                    & 5.27                    & 6.88                    & \textbf{42.83}      & \textbf{75.17}      & \textbf{56.41}      & 77.37               & \textbf{69.61}      \\
QuIP\#               & 4                    & \textbf{5.56}           & \textbf{7.07}           & \textbf{5.19}           & \textbf{6.75}           & 40.5                & 69.1                & ---                 & \textbf{78.4}       & 67.6                \\
AQLM                 & 4.04                 & ---                     & ---                     & 5.21                    & \textbf{6.75}           & 41.0                & 70.2                & 56.0                & 78.2                & 67.3                \\
VPTQ                 & 4.01                 & 5.64                    & 7.13                    & 5.26                    & 6.8                     & 39.7                & 69.0                & 56.0                & 78.1                & 67.1                \\ \hline
GPTQ                 & 3                    & ---           & ---            & 8.06                    & 10.61                   & 31.1                & 58.5                & 45.2                & 71.5                & 59.2                \\
GPTVQ                & 3.125                & 5.83                    & 7.51                    & 5.44                    & 7.24                    & \textbf{39.93}      & \textbf{74.07}      & 54.21               & 76.17               & \textbf{69.06}      \\
QuIP\#               & 3                    & \textbf{5.79}           & \textbf{7.32}           & \textbf{5.41}           & \textbf{7.04}           & 39.2                & 68.4                & ---                   & \textbf{77.3}       & 66.5                \\
AQLM                 & 3.04                 & ---                     & ---                     & 5.46                    & 7.08                    & 38.4                & 68.1                & 54.1                & 76.9                & 66.9                \\
VPTQ                 & 3.02                 & 5.82                    & 7.33                    & 5.43                    & \textbf{7.04}           & 39.3                & 69.1                & \textbf{54.9}       & \textbf{77.3}       & 68.0                \\ \hline
                     &                      &                         &                         &                         &                         &                     &                     &                     &                     &                     \\
\multirow{2}{*}{13B} & \multirow{2}{*}{bit} & \multirow{2}{*}{W2(2k)} & \multirow{2}{*}{C4(2k)} & \multirow{2}{*}{W2(4k)} & \multirow{2}{*}{C4(4k)} & \multirow{2}{*}{AC} & \multirow{2}{*}{AE} & \multirow{2}{*}{HE} & \multirow{2}{*}{QA} & \multirow{2}{*}{WI} \\
                     &                      &                         &                         &                         &                         &                     &                     &                     &                     &                     \\ \hline
GPTQ                 & 4                    & ---                     & ---                     & 4.78                    & 6.34                    & 42.49               & 70.45               & 58.67               & 77.75               & \textbf{70.01}      \\
GPTVQ                & 4.125                & 5.68                    & 7.25                    & 5.27                    & 6.88                    & 42.83               & \textbf{75.17}      & 56.41               & 77.37               & 69.61               \\
QuIP\#               & 4                    & \textbf{4.95}           & \textbf{6.54}           & \textbf{4.63}           & \textbf{6.13}           & \textbf{45.50}      & 73.90               & ---                 & \textbf{78.90}      & 69.90               \\
AQLM                 & 3.94                 & ---                     & ---                     & 4.65                    & 6.14                    & 44.80               & 73.32               & 59.27               & 78.35               & 69.85               \\
VPTQ                 & 4.02                 & 4.96                    & \textbf{6.54}           & 4.64                    & \textbf{6.13}           & 44.37               & 73.19               & \textbf{59.37}      & 77.75               & 69.77               \\ \hline
GPTQ                 & 3                    & ---                     & ---                     & 5.85                    & 7.86                    & 38.48               & 65.66               & 53.47               & 76.50               & 63.93               \\
GPTVQ                & 3.125                & 5.11                    & 6.83                    & 4.8                     & 6.47                    & \textbf{44.45}      & \textbf{77.23}      & 58.18               & 77.8                & \textbf{71.98}      \\
QuIP\#               & 3                    & \textbf{5.1}            & 6.72                    & \textbf{4.78}           & 6.35                    & 44.00               & 72.50               & ---                 & \textbf{78.40}      & 69.10               \\
AQLM                 & 3.03                 & ---                     & ---                     & 4.82                    & 6.37                    & 42.58               & 70.88               & 58.30               & 77.26               & 68.43               \\
VPTQ                 & 3.03                 & 5.12                    & \textbf{6.7}            & 4.79                    & \textbf{6.32}           & 42.32               & 73.99               & \textbf{58.42}      & 77.64               & 68.67               \\ \hline
                     &                      &                         &                         &                         &                         &                     &                     &                     &                     &                     \\
\multirow{2}{*}{70B} & \multirow{2}{*}{bit} & \multirow{2}{*}{W2(2k)} & \multirow{2}{*}{C4(2k)} & \multirow{2}{*}{W2(4k)} & \multirow{2}{*}{C4(4k)} & \multirow{2}{*}{AC} & \multirow{2}{*}{AE} & \multirow{2}{*}{HE} & \multirow{2}{*}{QA} & \multirow{2}{*}{WI} \\
                     &                      &                         &                         &                         &                         &                     &                     &                     &                     &                     \\ \hline
GPTQ                 & 4                    & ---                     & ---                     & 3.35                    & 5.15                    & 49.15               & 76.81               & 63.47               & 81.23               & 75.61               \\
GPTVQ                & 4.125                & 5.32                    & ---                     & ---                     & ---                     & ---                 & ---                 & ---                 & ---                 & ---                 \\
QuIP\#               & 4                    & \textbf{3.38}           & \textbf{5.56}           & \textbf{3.18}           & \textbf{5.02}           & 50.6                & 78.1                & ---                   & 81.4                & \textbf{77.1}       \\
AQLM                 & 4.14                 & ---                     & ---                     & 3.19                    & 5.03                    & \textbf{50.68}      & 77.31               & 63.69               & \textbf{81.5}       & 76.48               \\
VPTQ                 & 4.01                 & 3.39                    & 5.57                    & 3.19                    & \textbf{5.02}           & 49.57               & \textbf{78.16}      & \textbf{63.71}      & 81.18               & 76.4                \\ \hline
GPTQ                 & 3                    & ---                     & ---                     & 4.4                     & 6.26                    & 44.11               & 72.73               & 60                  & 78.4                & 71.82               \\
GPTVQ                & 3.125                & 5.51                    & ---                     & ---                     & ---                     & ---                 & ---                 & ---                 & ---                 & ---                 \\
QuIP\#               & 3                    & 3.56                    & \textbf{5.67}           & 3.35                    & \textbf{5.15}           & \textbf{50.9}       & \textbf{77.7}       & ---                 & \textbf{81.4}       & 76.4                \\
AQLM                 & 3.01                 & ---                     & ---                     & 3.36                    & 5.17                    & 50                  & 77.61               & 63.23               & 81.28               & 77.19               \\
VPTQ                 & 3.01                 & \textbf{3.55}           & \textbf{5.67}           & \textbf{3.34}           & \textbf{5.15}           & 48.89               & 77.06               & \textbf{63.52}      & 80.9                & \textbf{77.51}      \\ \hline
\end{tabular}
}
\end{table*}
\begin{table*}[hbt!]
\centering
\caption{LLaMA-3 Wikitext2 perplexity (context length 2048) and zeroshot QA Accuracy.}
\label{tab:llama3-234bit}
\resizebox{0.94\textwidth}{!}{%
\begin{tabular}{c|ccccccc|ccccccc}
\hline
\multirow{4}{*}{} & \multicolumn{7}{c|}{\multirow{2}{*}{LLaMA-3 8B}}                                                                                                                & \multicolumn{7}{c}{\multirow{2}{*}{LLaMA-3 70B}}                                                                                                                \\
                  & \multicolumn{7}{c|}{}                                                                                                                                          & \multicolumn{7}{c}{}                                                                                                                                           \\ \cline{2-15} 
                  & \multirow{2}{*}{bit} & \multirow{2}{*}{W2↓} & \multirow{2}{*}{AC↑} & \multirow{2}{*}{AE↑} & \multirow{2}{*}{HE↑} & \multirow{2}{*}{QA↑} & \multirow{2}{*}{WI↑} & \multirow{2}{*}{bit} & \multirow{2}{*}{W2↓} & \multirow{2}{*}{AC↑} & \multirow{2}{*}{AE↑} & \multirow{2}{*}{HE↑} & \multirow{2}{*}{QA↑} & \multirow{2}{*}{WI↑} \\
                  &                      &                      &                      &                      &                      &                      &                      &                      &                      &                      &                      &                      &                      &                      \\ \hline
FP16          & 16                   & 6.14                 & 50.3                 & 80.1                 & 60.2                 & 79.6                 & 73.1                 & 16                   & 2.9                  & 60.1                 & 87.0                 & 66.3                 & 82.4                 & 80.8                 \\
QuIP              & 4                    & 6.5                  & 47.4                 & 78.2                 & 58.6                 & 78.2                 & 73.2                 & 4                    & 3.4                  & 58.7                 & 86.0                 & 65.7                 & 82.5                 & 79.7                 \\
GPTQ              & 4                    & 6.5                  & 47.7                 & 78.8                 & 59.0                 & 78.4                 & 72.6                 & 4                    & 3.3                  & 58.4                 & \textbf{86.3}                 & 66.1                 & \textbf{82.9}                 & \textbf{80.7}                 \\
VPTQ              & 4.03                 & \textbf{6.42}        & \textbf{49.1}        & \textbf{78.8}        & \textbf{59.3}        & \textbf{78.7}        & \textbf{74.8}        & 4.05                 & \textbf{3.15}        & \textbf{59.0}        & 86.1                 & \textbf{66.2}        & 82.4                 & 79.8                 \\ \hline
QuIP              & 3                    & 7.5                  & 41.0                 & 72.9                 & 55.4                 & 76.8                 & 72.5                 & 3                    & 4.7                  & 54.9                 & 83.3                 & 63.9                 & \textbf{82.3}                 & 78.4                 \\
GPTQ              & 3                    & 8.2                  & 37.7                 & 70.5                 & 54.3                 & 74.9                 & 71.1                 & 3                    & 5.2                  & 52.1                 & 79.6                 & 63.5                 & 80.6                 & 77.1                 \\
VPTQ              & 3.03                 & \textbf{6.97}        & \textbf{45.8}        & \textbf{77.5}        & \textbf{58.4}        & \textbf{78.2}        & \textbf{73.4}        & 3.01                 & \textbf{3.81}        & \textbf{57.3}        & \textbf{84.7}        & \textbf{65.5}        & 81.7                 & \textbf{79.2}        \\ \hline
QuIP              & 2                    & 85.1                 & 21.3                 & 29.0                 & 29.2                 & 52.9                 & 51.7                 & 2                    & 13                   & 26.5                 & 48.9                 & 40.9                 & 65.3                 & 61.7                 \\
DB-LLM            & 2                    & 13.6                 & 28.2                 & 59.1                 & 42.1                 & 68.9                 & 60.4                 & N/A                    & N/A                   & N/A                    & N/A                     & N/A                  & N/A                     & N/A                      \\
GPTQ              & 2                    & 2.10E+02             & 19.9                 & 28.8                 & 27.7                 & 53.9                 & 50.5                 & 2                    & 11.9                 & 24.6                 & 38.9                 & 41.0                 & 62.7                 & 59.9                 \\
VPTQ              & 2.08                 & \textbf{9.29}        & \textbf{36.9}        & \textbf{71.0}        & \textbf{52.2}        & \textbf{75.1}        & \textbf{65.9}        & 2.02                 & \textbf{5.6}         & \textbf{52.5}        & \textbf{81.8}        & \textbf{61.7}        & \textbf{80.4}        & \textbf{77.9}        \\
VPTQ              & 2.24                 & \textbf{9.19}        & \textbf{42.6}        & \textbf{73.2}        & \textbf{53.1}        & \textbf{75.4}        & \textbf{69.1}        & 2.07                 & \textbf{5.66}        & \textbf{54.2}        & \textbf{83.6}        & \textbf{61.8}        & \textbf{80.1}        & \textbf{74.0}        \\ \hline
\end{tabular}%
}
\end{table*}
\begin{table*}[hbt!]
\centering
\caption{Mistral-7B-v0.1 Wikitext2, C4 perplexity (context length 2048 and 8192) and zeroshot QA Accuracy}
\label{tab:mistral-7b-234bit}
\resizebox{0.8\textwidth}{!}{%
\begin{tabular}{c|ccccccccc}
\hline
                  & \multicolumn{9}{c}{Mistral 7b}                                                                                                                                                                                   \\ \hline
\multirow{2}{*}{} & \multirow{2}{*}{bit} & \multirow{2}{*}{W2(2k)} & \multirow{2}{*}{W2(8k)} & \multirow{2}{*}{C4(8k)} & \multirow{2}{*}{AC} & \multirow{2}{*}{AE} & \multirow{2}{*}{HE} & \multirow{2}{*}{QA} & \multirow{2}{*}{WI} \\
                  &                      &                         &                         &                         &                     &                     &                     &                     &                     \\ \hline
FP16              & 16                   & 5.25                    & 4.77                    & 5.71                    & 48.89               & 78.87               & 61.12               & 80.3                & 73.88               \\
GPTVQ             & 4.125                & 5.38                    & 4.87                    & 6.13                    & \textbf{50}         & \textbf{80.43}      & 60.36               & 79.65               & 73.4                \\
QuIP\#            & 4                    & ---                     & 4.85                    & 5.79                    & 49.4                & 78.96               & 60.62               & \textbf{80.41}      & 73.95               \\
AQLM              & 4.02                 & ---                     & 4.85                    & 5.79                    & 48.21               & 77.86               & 60.27               & 79.71               & 73.8                \\
GPTQ              & 4.125                & \textbf{5.36}           & 4.83                    & 5.74                    & 49.57               & 79.5                & 60.38               & 79.54               & 72.85               \\
VPTQ              & 4.03                 & \textbf{5.36}           & \textbf{4.81}           & \textbf{5.72}           & 48.12               & 77.82               & \textbf{60.61}      & 80.14               & \textbf{74.19}      \\ \hline
GPTVQ             & 3.125                & 6.42                    & 6.8                     & 13.28                   & 40.78               & 75.67               & 54.18               & 77.42               & 67.4                \\
AQLM              & 3.04                 & ---                     & 5.07                    & 5.97                    & 46.67               & 77.61               & 59.31               & \textbf{80.14}      & \textbf{72.69}      \\
GPTQ              & 3.125                & 6.02                    & 5.88                    & 6.86                    & \textbf{47.35}      & 77.86               & 58.84               & 79.82               & 71.74               \\
VPTQ              & 3.03                 & \textbf{5.53}           & \textbf{4.96}           & \textbf{5.84}           & 46.67               & \textbf{77.95}      & \textbf{59.91}      & 79.49               & 72.45               \\ \hline
QuIP\#            & 2                    & ---                     & 6.02                    & 6.84                    & 39.76               & 72.14               & 52.95               & 76.71               & \textbf{69.3}       \\
AQLM              & 2.01                 & ---                     & 6.32                    & 6.93                    & 40.44               & \textbf{73.65}      & 52.13               & 76.01               & 68.75               \\
GPTVQ             & 2.25                 & 8.2                     & 8.99                    & 18.6                    & 37.37               & 71                  & 45.43               & 70.18               & 64.33               \\
GPTQ              & 2.125                & 280                     & 1535                    & 164                     & 24.49               & 44.91               & 36.56               & 63.33               & 52.96               \\
VPTQ              & 2.04                 & \textbf{6.32}           & \textbf{5.64}           & \textbf{6.43}           & \textbf{41.13}      & 72.22               & \textbf{56.1}       & \textbf{77.91}      & 68.67               \\ \hline
\end{tabular}
}
\end{table*}

\section{Quantitative Analysis of Quantization Parameter Settings}
\label{appendix:analysis}
\textbf{Quantization configuration}
The quantization parameters of all VPTQ 2bit models are shown in Table \ref{tab:quant_args}.
\begin{table*}[hbt!]

\caption{Parameters for 2-bit Quantization of Llama and Mistral Models. $v$ represents the vector length, $k$ denotes the codebook size, $k1$ and $k2$ correspond to the two codebooks, and $group\quad num$ indicates the number of groups into which PQ (Product Quantization) is divided.}
\label{tab:quant_args}
\centering\resizebox{0.62\textwidth}{!}
{
\begin{tabular}{cc|ccc|cccc}
\hline
                            & \multirow{2}{*}{bit} & \multicolumn{3}{c|}{Outlier} & \multicolumn{4}{c}{Other}    \\
                            &                      & N\%      & v      & k        & v  & k1   & k2   & group num \\ \hline
\multirow{2}{*}{LLaMA2-7b}  & 2.02                 & 0        & -      & -        & 6  & 4096 & -    & 1         \\
                            & 2.26                 & 1        & 4      & 8192     & 12 & 4096 & 4096 & 4         \\ \hline
\multirow{2}{*}{LLaMA2-13b} & 2.02                 & 0        & -      & -        & 6  & 4096 & -    & 1         \\
                            & 2.18                 & 2        & 4      & 8192     & 12 & 4096 & 4096 & 4         \\ \hline
\multirow{2}{*}{LLaMA2-70b}                 & 2.07                 & 1        & 4      & 8192     & 12 & 4096 & 4096 & 4         \\
                            & 2.11                 & 1        & 4      & 8192     & 12 & 4096 & 4096 & 8         \\ \hline
\multirow{2}{*}{LLaMA3-8b}  & 2.08                 & 1        & 4      & 4096     & 12 & 4096 & 4096 & 1         \\
                            & 2.24                 & 1        & 4      & 8192     & 6  & 4096 & -    & 16        \\ \hline
\multirow{2}{*}{LLaMA3-70b} & 2.02                 & 0        & -      & -        & 12 & 4096 & 4096 & 1         \\
                            & 2.07                 & 1        & 4      & 4096     & 6  & 4096 & -    & 16        \\ \hline
\end{tabular}%
}
\end{table*}
\begin{table}[hbt!]
\centering
\caption{Layer-wise finetuning parameters on 8xH100}
\label{tab:layer_ft_param}
\resizebox{0.8\columnwidth}{!}{%
\begin{tabular}{c|cc}
\hline
model       & finetune lr & batchsize \\ \hline
LLaMA-2-7B  & $1\times 10^{-4}$    & 32        \\
LLaMA-2-13B & $1\times 10^{-4}$    & 32        \\
LLaMA-2-70B & $1\times 10^{-5}$    & 16        \\
LLaMA-3-8B  & $1\times 10^{-5}$    & 16        \\
LLaMA-3-70B & $5\times 10^{-6}$    & 8         \\
Mistral-7B  & $5\times 10^{-6}$    & 16        \\ \hline
\end{tabular}%
}
\end{table}

\textbf{Layer-wise fine-tuning parameters}
Layer-wise finetuning trains centroids and layer norm using the input and output of each layer when entering 128 samples of C4 training sets into the full precision model. We train each layer for 100 iterations. Table \ref{tab:layer_ft_param} shows the learning rate and batch size used for each model.

\section{Ablation Study}
\label{sec:ablation_study}
\begin{table*}[hbt!]
\centering
\caption{Ablation Study on Different Quantization Techniques for LLaMA-2 13B}
\resizebox{0.9\textwidth}{!}{%
\begin{tabular}{c|c|c|cc|ccc|cccc|cc}
\hline
 &
  \multirow{2}{*}{bit} &
  \multirow{2}{*}{\begin{tabular}[c]{@{}c@{}}channel\\ independent\end{tabular}} &
  \multicolumn{2}{c|}{Finetune} &
  \multicolumn{3}{c|}{outlier} &
  \multicolumn{4}{c|}{other} &
   &
   \\ \cline{4-14} 
 &
   &
   &
  \begin{tabular}[c]{@{}c@{}}layer\\ wise\end{tabular} &
  e2e &
  N\% &
  v0 &
  k0 &
  v1 &
  k1 &
  k2 &
  \begin{tabular}[c]{@{}c@{}}group\\ num\end{tabular} &
  W2(↓) &
  C4(↓) \\ \hline
\#1  & FP16 & -   & -   & -   & - & - & -    & -  & -     & -    & - & 4.57  & 6.05  \\
\#2  & 2    & Yes & No  & No  & 0 & - & -    & 2  & 16    & -1   & 1 & 14800 & 13337 \\
\#3  & 2.01 & Yes & No  & No  & 0 & - & -    & 4  & 256   & -1   & 1 & 7.21  & 9.78  \\
\#4  & 2.02 & Yes & No  & No  & 0 & - & -    & 6  & 4096  & -1   & 1 & 6.29  & 8.29  \\
\#5  & 2.02 & No  & No  & No  & 0 & - & -    & 6  & 4096  & -1   & 1 & 7.25  & 9.8   \\
\#6  & 2.19 & Yes & No  & No  & 0 & - & -    & 8  & 65536 & -1   & 1 & 5.8   & 7.68  \\
\#7  & 2.04 & Yes & No  & No  & 0 & - & -    & 12 & 4096  & 4096 & 1 & 6.32  & 8.29  \\
\#8  & 2.03 & Yes & No  & No  & 1 & 4 & 4096 & 6  & 4096  & -1   & 1 & 6.16  & 8.08  \\
\#9  & 2.04 & Yes & No  & No  & 2 & 4 & 4096 & 6  & 4096  & -1   & 1 & 6.08  & 8.12  \\
\#10 & 2.07 & Yes & No  & No  & 5 & 4 & 4096 & 6  & 4096  & -1   & 1 & 6.02  & 7.96  \\
\#11 & 2.02 & Yes & Yes & No  & 0 & - & -    & 6  & 4096  & -1   & 1 & 6.07  & 7.64  \\
\#12 & 2.02 & Yes & Yes & Yes & 0 & - & -    & 6  & 4096  & -1   & 1 & 5.32  & 7.15  \\
\#13 & 2.04 & Yes & Yes & No  & 0 & - & -    & 12 & 4096  & 4096 & 1 & 5.71  & 7.52  \\
\#14 & 2.06 & Yes & Yes & No  & 1 & 4 & 4096 & 12 & 4096  & 4096 & 1 & 5.63  & 7.45  \\
\#15 & 2.09 & Yes & Yes & No  & 1 & 4 & 4096 & 12 & 4096  & 4096 & 2 & 5.63  & 7.41  \\
\#16 & 2.17 & Yes & Yes & No  & 1 & 4 & 4096 & 12 & 4096  & 4096 & 4 & 5.63  & 7.38  \\
\#17 & 2.3  & Yes & Yes & No  & 1 & 4 & 4096 & 12 & 4096  & 4096 & 8 & 5.55  & 7.38  \\
\#18 & 3.01 & Yes & Yes & No  & 0 & - & -    & 4  & 4096  & -1   & 1 & 4.82  & 6.37  \\
\#19 & 4.02 & Yes & Yes & No  & 0 & - & -    & 6  & 4096  & 4096 & 1 & 4.64  & 6.13  \\ \hline
\end{tabular}%
}

\label{tab:ablation_study}
\end{table*}
\begin{table}[hbt!]
\centering
\caption{Ablation of Vector Length on Inference Throughput and Peak Memory Usage}
\label{tab:ablation_performance}
\resizebox{\columnwidth}{!}{%
\begin{tabular}{c|cccc|cc}
\hline
     & v1 & k1    & k2   & group num & tok/s & mem(GB) \\ \hline
FP16 & -  & -     & -    & -          & 30.03              & 63.63                  \\
2    & 2  & 16    & -1   & 1          & 18.85              & 4.17                   \\
2.01 & 4  & 256   & -1   & 1          & 17.06              & 4                      \\
2.02 & 6  & 4096  & -1   & 1          & 32.09              & 4.02                   \\
2.19 & 8  & 65536 & -1   & 1          & 30.64              & 4.46                   \\
2.04 & 12 & 4096  & 4096 & 1          & 21.34              & 4.06                   \\ \hline
\end{tabular}}
\end{table}
Table \ref{tab:ablation_study} shows results from LLaMA-2 13b on Wikitext2 and C4 (sequence length = 4096) under different quantization parameters. The impact of techniques such as vector length, channel-independent optimization, residual vector quantization, outlier elimination, layer-wise fine-tuning, and end-to-end fine-tuning on quantization results will be discussed.
\subsection{Parameter Description}
When performing N\% outlier elimination, N\% of outliers will be quantized using a codebook with a vector length of $v_0$ and $k_0$ centroids. For the remaining (100-N)\% parameters, the vector length is $v_1$. $k_1$ represents the number of centroids in the first codebook, while $k_2$ represents the number of centroids in the second codebook for residual vector quantization. $k_2=-1$ indicates no residual vector quantization.
\subsection{Vector Length and Residual Vector Quantization}
\textbf{Compression Ratio Calculation} 
The average bitwidth per element of the index matrix obtained through vector quantization is:

\begin{equation}
    \text{Average index bitwidth}=\frac{\log_2(k_1)}{v_1} + \frac{\log_2(k_2)}{v_1}
    \nonumber
\end{equation}

The compression ratio is calculated by:
\begin{equation}
    \text{Compression ratio} = \frac{\text{Total original model bits}}{\text{Codebook bits} + \text{Index bits}}
    \nonumber
\end{equation}

For an original linear weight matrix with $M$ parameters,
\begin{equation}
\text{Codebook bits} = (v_0 \times k_0
    + v_1 \times (k_1 + k_2))  \times 16
    \nonumber
\end{equation}

\begin{equation}
\begin{aligned}
    \text{Index bits} & =  M \times N\% \times \log_2\left(\frac{k_0}{v_0}\right)  + M \times \\ & (100-N)\% \times \left[ \frac{\log_2(k_1)}{v_1} + \frac{\log_2(k_2)}{v_1} \right]
    \nonumber
\end{aligned}
\end{equation}

The total bitwidth in the table is calculated per transformer block, which for LLaMA-2 includes 4 attention linear and 3 FFN linear layers.

\textbf{Impact of Vector Length} First, we discuss the impact of vector length on accuracy. In Table \ref{tab:ablation_study} rows \#2, \#3, \#4, and \#6 show results for $v_1=2, 4, 6, 8$, keeping the average index bit at 2 (i.e., $log_2(k_1/v_1) = 2$). As $v_1$ increases, the perplexity on Wikitext2 and C4 decreases, but the codebook size also increases exponentially. For $v_1=8$ and $k_1=65536$, the codebook overhead introduces an additional 0.19 bits. 
Then, we evaluate the model inference throughput in Table \ref{tab:ablation_performance}. Since we employ weight-only quantization, the main additional overhead of quantized model inference comes from the lookup table for model weights. Table \ref{tab:ablation_performance} shows models with ~2 bits on various throughputs. As the vector length increases (from 2 to 6), the granularity of memory access for reading the lookup table in dequantization increases, which allows memory access to match the GPU's cache line (128 bytes @ L1). This reduces memory access transactions and decreases cache misses. As the vector length further increases (from 8 to 12) along with the size and levels of the codebook, the codebook size further increases, which results in the codebook not fitting in the L1 cache, thereby reducing the model's inference speed. Additionally, we find that a reasonable setting (e.g., \(v=6\), \(k=4096\)) can achieve throughput similar to the original model for the quantized model, demonstrating the efficiency of the VPTQ design.

\textbf{Residual Vector Quantization} Without any fine-tuning, rows \#4 and \#7 show similar perplexities for $v_1=6, k_1=4096$ and $v_1=12, k_1=k_2=4096$ , with the latter even higher. However, after layer-wise fine-tuning, comparing rows \#11 and \#13, residual vector quantization (RVQ) reduces the perplexity by 0.3 compared to vector quantization (VQ) due to the increased number of finetunable centroids, showing significant improvement.

\subsection{Channel-Independent Optimization}
Row \#4 with channel-independent optimization shows a perplexity decrease of 1 compared to row \#5 without it, indicating that channel-independent second-order optimization effectively mitigates quantization error accumulation.

\subsection{Outlier Elimination}
Rows \#4, \#8, \#9, and \#10 represent the results for eliminating 0\%, 1\%, 2\%, and 5\% outliers, respectively. We used a codebook with $v_0=4$ and $k_0=4096$ to quantize N\% of outliers, achieving an effective average index bit of 3 bits, while other parameters were 2 bits. Higher N\% means more parameters are quantized with 3 bits, leading to a larger total bitwidth and lower perplexity.

\subsection{Fine-tuning}
Rows \#4, \#11, and \#12 show results without any fine-tuning, with layer-wise fine-tuning, and with end-to-end fine-tuning, respectively. Adding fine-tuning reduced the perplexity on Wikitext2 from 6.29 to 6.07 and further to 5.32.

\subsection{Group Number}
Rows \#14, \#15, \#16, and \#17 show the quantization results when 99\% of parameters are divided into 1, 2, 4, and 8 groups, respectively. Each group has its own independent codebook. When divided into 1, 2, and 4 groups, the perplexity on Wikitext2 does not change much, likely because the distribution of the remaining parameters (after removing 1\% outliers) is relatively uniform. This is likely because the distributions of different groups overlap after grouping, so the benefit of increasing the group number is not significant.

\subsection{Higher Bitwidth}
Rows \#18 and \#19 represent the results for 3-bit and 4-bit quantization, respectively. Compared to the FP16 results in row \#1, 4-bit vector quantization incurs almost no loss.

\section{Inference Evaluation}
\subsection{Throughput Measurement Process}
We follow the throughput measurement method used in AQLM \cite{AQLM}. During the prompt phase, we provide 1 token and then have the model generate 256 tokens, calculating the generation time for each output token to determine the throughput in tokens per second (tok/s).

\subsection{Our Dequantization Implementation}
Our dequantization implementation is divided into two phases. In the first phase, which handles prompts with relatively long sequences, we restore the quantized weights (index and centroid, etc.) to FP16 and then call `torch.matmul`.  In the second phase, during decoding, we fuse the dequantization and GEMV operations into QGemv, eliminating the repetitive reading and writing of FP16 weights.

\end{document}